\documentclass[11pt]{article}
\usepackage[T1]{fontenc}
\usepackage[utf8]{inputenc}
\usepackage{lmodern}
\usepackage{amsmath,amssymb}
\usepackage{graphicx}
\usepackage{hyperref}
\usepackage{booktabs}
\usepackage{array}
\usepackage{calc}

\title{Lossless Compression of Neural Network Components: Weights, Checkpoints, and K/V Caches in Low-Precision Formats}
\author{
  Anat Heilper \\
  \small Intel \\
  \small \texttt{anat.heilper@intel.com}
  \and
  Doron Singer \\
  \small Intel \\
  \small \texttt{dsinger@intel.com}
}
\date{August 11, 2025}

\begin{document}
\maketitle

\begin{abstract}
As deep learning models grow and deployment becomes more widespread, 
reducing the storage and transmission costs of neural network weights 
has become increasingly important. While prior work such as ZipNN~\cite{zipnn} 
has shown that lossless compression methods—particularly those based 
on Huffman~\cite{huffman1952} encoding floating-point exponents can 
significantly reduce model sizes, these techniques have primarily been 
applied to higher-precision formats such as FP32 and BF16. In this work, 
we extend the ZipNN~\cite{zipnn} approach to lower-precision floating-point formats, 
specifically FP8 and FP4, which are gaining popularity for efficient 
inference. We design a compression method that separates and compresses 
the exponent and mantissa components independently using entropy coding. 
Our evaluation shows compression ratios up to 62\% for BF16 and 83\% for 
FP8. We also investigate the compressibility of key-value (K/V) cache 
tensors used in large language models (LLMs), finding that they, too, 
exhibit compressible patterns, enabling memory savings during deployment.
\end{abstract}

\section{Introduction}

The rapid advancement of deep learning has led to increasingly large
models, which require significant storage, memory, and communication
resources. With the growing demand for deploying these models in various
settings, including edge devices, data centers, and model hubs such as
HuggingFace~\cite{huggingface_transformers}, there is a strong need for efficient mechanisms to
store and transfer model parameters. Although techniques such as
quantization, pruning, and knowledge distillation reduce model sizes by
sacrificing some precision or structure, there remains a need for
\emph{lossless} methods that preserve the original model exactly while
still achieving compression.

Recent work by Hershcovitch et al. introduced ZipNN~\cite{zipnn}, a lossless
compression framework tailored for deep learning models. ZipNN~\cite{zipnn}
identifies a key insight: the exponent part of floating-point weights,
particularly in formats like BF16, exhibits a highly skewed distribution
that can be effectively compressed using entropy-based techniques such
as Huffman~\cite{huffman1952} coding. By separating the exponent from the mantissa
and applying targeted encoding, ZipNN~\cite{zipnn} achieves impressive
compression ratios, with additional speedups in both compression and
decompression. However, while ZipNN~\cite{zipnn} focuses on compressing
weights and checkpoints, it does not address the key-value (k/v) cache,
which represents a significant storage consumer during inference,
particularly for transformer-based models.

ZipNN~\cite{zipnn} primarily addresses higher-precision formats like FP32,
FP16, and BF16. As the industry shifts toward lower-precision formats
such as FP8 and FP4 to reduce computational cost and memory bandwidth
during inference, it remains unclear whether similar compression
benefits can be achieved at such low bit widths. FP8 and FP4 offer
limited representation capacity, and it is not obvious that the exponent
in these formats will still exhibit compressible statistical patterns.

In this work, we investigate whether the ZipNN~\cite{zipnn} methodology can be
extended to FP8 and FP4 weights. We analyze the distributions of
exponent and mantissa values across various models and training
checkpoints and design a compression scheme that separately encodes
these components using Huffman~\cite{huffman1952} coding. Our results show that
even in these ultra-low-precision formats, the exponent distributions
are skewed enough to allow for significant compression, while the
mantissa compresses less effectively. We report exponent compression
ratios as low as 0.07 in checkpoint compression use case  and total model compression down to 37 percent of
the original size in some cases.

Our contributions can be summarized as follows:

\begin{itemize}
\item
  We extend Huffman~\cite{huffman1952}-based lossless compression techniques to FP8
  and FP4 formats.
\item
  We show that even low-bit exponents retain skewed distributions,
  enabling effective entropy encoding.
\item
  We evaluate our approach across multiple training checkpoints,
  demonstrating consistent compression gains.
\item
  We provide an analysis of how compression effectiveness varies across
  training stages and tensor components.
\end{itemize}

The remainder of the paper is organized as follows. Section 2 provides
background on floating-point formats and related work in model
compression. Section 3 describes our proposed methodologies. Section 4 presents
our experimental results. Section 5 discusses the implications of our
findings and future directions, and Section 6 concludes the paper.

\section{Background and Related
Work}\label{background-and-related-work}

\subsection{Floating Point Formats in Deep
Learning}\label{floating-point-formats-in-deep-learning}

Modern neural networks typically represent model parameters and
intermediate tensors using floating-point formats. While early
architecture relied heavily on 32-bit floating point (FP32), there has
been a steady shift toward reduced-precision formats to improve memory
efficiency and computational throughput. BF16 and FP16 are now widely
used during training and inference due to their favorable trade-offs
between dynamic range and precision. More recently, formats such as FP8
and FP4 have gained attention as hardware vendors introduce support for
ultra-low-precision inference. These formats typically use fewer bits
for the exponent and mantissa, for example FP8 variants like E4M3 or
E5M2, which allow for a dynamic range sufficient for many inference
tasks while minimizing bandwidth and storage.

\subsection{Lossless Compression of Neural Network
Weights}\label{lossless-compression-of-neural-network-weights}

Traditional model compression methods include quantization, pruning, and
knowledge distillation. These techniques reduce model size by modifying
its structure or representation, often introducing minor losses in
accuracy. In contrast, lossless compression techniques aim to reduce
size while preserving exact numerical equivalence after decompression.

The ZipNN~\cite{zipnn} framework introduced a specialized lossless 
compression approach tailored for deep learning models. 
ZipNN~\cite{zipnn} observed that in floating-point representations,
the exponent bits are often highly skewed and thus compressible using
entropy coding such as Huffman~\cite{huffman1952} encoding. By separating the
exponent from the mantissa and applying Huffman~\cite{huffman1952} coding
exclusively to the exponent, ZipNN~\cite{zipnn} achieved substantial
compression for BF16 and FP32 models, particularly those with clean or
sparsified weights. This method also avoided the inefficiencies of
general-purpose Lempel-Ziv (LZ) algorithms, which are not well-suited
for floating-point tensors with limited multi-byte repetition.

In this work, we extend the ZipNN~\cite{zipnn} strategy to low-precision
formats like FP8 and FP4. These formats use smaller exponents, which
makes the entropy distribution even more critical for compression.
Despite their compactness, we find that the exponent values still
exhibit sufficient redundancy to be effectively compressed.

\subsection{Related Work in Compression for Floating-Point
Data}\label{related-work-in-compression-for-floating-point-data}

Beyond ZipNN~\cite{zipnn}, there has been prior work in compressing
floating-point data for scientific computing and machine learning. For
instance, the ZFP library~\cite{zfp} introduces block-based
compression of floating-point arrays using a shared exponent within
blocks and bit-plane encoding of mantissas. Although ZFP~\cite{zfp} achieves good
ratios for scientific data, it is optimized for structured grid data and
typically operates in lossy mode, which is not suitable for preserving
exact model parameters.

Other approaches, such as dietgpu~\cite{dietgpu}, have implemented lightweight
entropy-based compression schemes for GPU-accelerated workloads.
DietGPU~\cite{dietgpu} includes a mode for exponent-only compression, similar in
spirit to ZipNN~\cite{zipnn}, although it does not target clean model
representations or analyze FP8/FP4 explicitly.

General-purpose lossless compressors like Zstd~\cite{zstd_spec} and Zlib~\cite{zlib_spec}
have also been applied to neural network weights with mixed results.
These tools are effective when applied to models with clear byte-level
regularities, but they are not optimized for floating-point data and
often fail to exploit the structure of exponent-mantissa encoding.

\subsection{Key-Value Cache in Large Language
Models}\label{key-value-cache-in-large-language-models}

An emerging area of interest is the memory overhead introduced by
\textbf{key-value (K/V) cache} tensors in large language models (LLMs).
During autoregressive decoding, LLMs must store hidden states from
previous tokens to compute future attention layers. These K/V caches
often grow linearly with sequence length and batch size, making them a
memory bottleneck during inference, especially in low-latency settings.
Recent efforts have focused on reducing K/V memory usage through
sparsity or quantization. However, to our knowledge, no prior work has
explored lossless compression of K/V caches.

In this work, we analyze the structure of FP8-formatted K/V tensors and
observe that their exponent distributions are often similarly skewed,
making them candidates for compression. We apply our
Huffman~\cite{huffman1952}-based encoding strategy to these caches and show that
they too benefit from exponent-mantissa separation, offering runtime
memory savings without modifying inference logic.

\section{Methodology}\label{methodology}

This section outlines the compression strategies we developed for
various low-precision formats and contexts. Our techniques are grounded
in entropy-based encoding, particularly Huffman~\cite{huffman1952} coding, and are
applied differently depending on the structure and statistical
properties of the data. Where appropriate, we adopt the separation of
exponent and mantissa components as proposed in the ZipNN~\cite{zipnn}
framework and extend it to new use cases, including FP8, FP4, delta
checkpoints, and K/V cache tensors.

\subsection{Compression of BF16 Delta
Checkpoints}\label{compression-of-bf16-delta-checkpoints}

Delta checkpoints represent the difference between two successive
versions of a model during training or fine-tuning. In our work, we
compress the delta between BF16 checkpoint files, which are commonly
used in distributed training. Each parameter is stored in 16-bit format
with a 1-bit sign, 8-bit exponent, and 7-bit mantissa.

We apply a block wise XOR operation between consecutive checkpoints to
compute the delta. The result often exhibits a higher density of zeros,
especially in early training stages or when fine-tuning has converged.
Following this step, we extract the exponent and mantissa bits from the
delta values and compress them independently. The exponent stream is
highly skewed, allowing efficient compression via Huffman~\cite{huffman1952}
coding. The mantissa stream is evaluated for entropy; if compressibility
is high, we apply Huffman~\cite{huffman1952} encoding, otherwise it is stored
uncompressed.

Compression is performed in fixed-size chunks with lightweight metadata
stored per block. These chunks are designed to support random access and
parallel decoding.

\subsection{Compression of FP8
Weights}\label{compression-of-fp8-weights}

FP8 formats are increasingly adopted for inference due to their
compactness and support from emerging hardware platforms. In our
experiments, we focus on two commonly used variants: E4M3 and E5M2.
These formats allocate 4 or 5 bits to the exponent and the remaining
bits to the mantissa, along with a single sign bit.

We process FP8 tensors by extracting the exponent and mantissa streams
and applying Huffman~\cite{huffman1952} coding separately to each. As in BF16, the
exponent distribution is typically skewed and thus compressible. Despite
the narrow dynamic range of FP8, the skew is often pronounced enough to
yield compression ratios significantly below 0.3 in some cases. The
mantissa stream is usually less compressible, although we observe some
redundancy in specific layers and at early training stages.

\includegraphics[width=3.63213in,height=1.29868in]{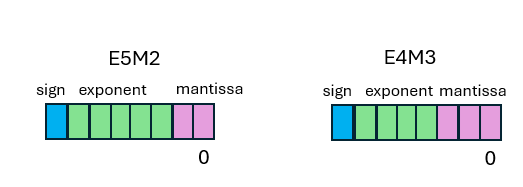}

\emph{Figure 1: FP8 formats E5m2, E4m3}

\subsection{Compression of K/V Cache Tensors (BF16 and
FP8)}\label{compression-of-kv-cache-tensors-bf16-and-fp8}

During inference in large language models, the key-value (K/V) cache
stores intermediate hidden states for each attention layer. These caches
accumulate across decoding steps and become a major source of memory
usage, particularly for long sequences. We examine K/V caches formatted
in BF16 and FP8, depending on model configuration.

We apply our compression technique to K/V cache tensors in a layer-wise
manner. Exponent and mantissa bits are separated, and Huffman~\cite{huffman1952}
coding is applied to the exponent stream. In FP8 caches, the exponent
remains skewed across time steps, especially in early tokens, enabling
compression. In BF16, the larger exponent width allows even greater
compressibility. Mantissa compressibility depends on the
model\textquotesingle s internal dynamics, but we observed that the
exponent stream consistently offers savings.

Since K/V caches are generated and accessed during runtime, we optimize
the compression process for low latency. In practice, we use precomputed
Huffman~\cite{huffman1952} dictionaries when exponent distributions are stable,
and we update them adaptively only when compression ratios drop.

This is, to the best of our knowledge, the first demonstration of
effective lossless compression of K/V cache tensors. The resulting
savings can help reduce memory usage and may allow longer sequences or
higher batch sizes in memory-constrained environments.

\subsection{Compression of FP4
Weights}\label{compression-of-fp4-weights}

FP4 formats are the most aggressive low-precision representations under
exploration for model deployment. Unlike FP8 or BF16, FP4 formats
typically represent values indirectly using scaling factors. Two main
variants are in use: \textbf{MXFP4~\cite{ocp_mx_format}} and \textbf{NVFP4~\cite{nvidia_nvfp4}}. 

In MXFP4~\cite{ocp_mx_format}, tensors are stored as 4-bit unsigned integers
In MXFP4~\cite{ocp_mx_format}, tensors are stored as 4-bit unsigned integers
accompanied by a shared floating-point scaling factor per block (usually
per row or group of rows).

The figure below shows the sign, exponent, and mantissa fields in an
MX-compliant format with floating-point element data type (left) and an
MX-compliant format with integer element data type (right).

\includegraphics[width=4.36134in,height=1.132in]{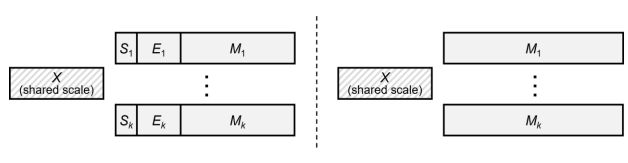}

\emph{Figure 2: shared scaling factor per group of elements in MX
compliant format. Taken from
\href{https://www.opencompute.org/documents/ocp-microscaling-formats-mx-v1-0-spec-final-pdf}{OCP
standard}}

\href{https://docs.nvidia.com/deeplearning/cudnn/frontend/v1.13.0/operations/BlockScaling.html}{NVFP4}~\cite{nvidia_nvfp4}:
The NVFP4~\cite{nvidia_nvfp4} recipe quantizes across 16 FP32 elements along the rows
to produce 16 FP4 output values (E2M1) and 1 FP8 scaling factor (E4M3).

\includegraphics[width=5.21555in,height=1.71537in]{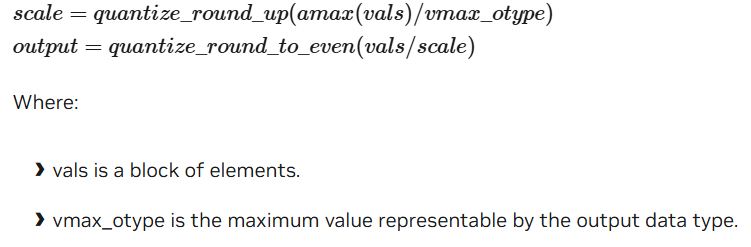}

\emph{Figure 3: NVFP4 conversion of FP4 elements with scaling factors}

For the context of this paper, we're interested in capacity perspective
only, the differences between these formats can be simplified to the
following table:

\includegraphics[width=5.21555in,height=1.71537in]{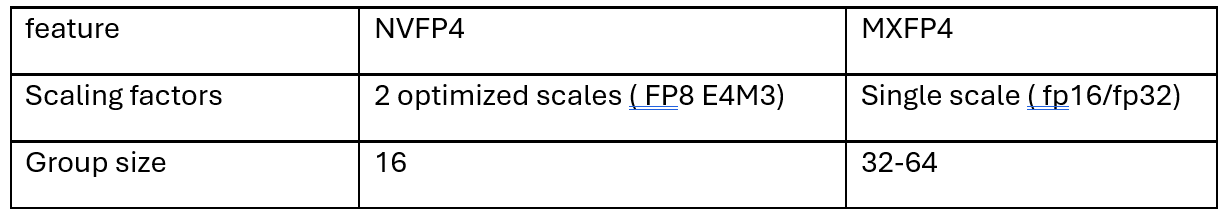}

\emph{Figure 4: Difference in the demand for scale factors capacity between NVFP4 and MXFP4 }

To investigate whether the 4-bit quantized values contain any
compressible structure, we conducted a series of experiments. One
approach involved extracting the exponent-like bits from multiple FP4
entries and combining them to form larger binary words. For instance, we
selected 2 bits from each of the 4 consecutive FP4 values to build an
8-bit stream. The motivation was to create a byte-aligned representation
that might contain repeated patterns across the model. However,
empirical results showed that these transformed streams did not yield
meaningful compression. The bit patterns appeared close to uniformly
random, and entropy-based encoding did not reduce the data size
significantly.

As a result, we conclude that the quantized values in FP4 do not exhibit
consistent statistical redundancy and are not suitable for lossless
compression. On the other hand, the scaling factors used in FP4 blocks
do show structure that can be compressed effectively. These factors
often follow smooth distributions or repeated patterns, especially in
transformer layers where normalization and activation scaling are
applied uniformly. Applying Huffman~\cite{huffman1952} coding to the scaling factor
stream leads to measurable compression gains with minimal processing
overhead.

In summary, our FP4 compression strategy targets only the scaling
factors and stores the quantized values uncompressed. This selective
approach provides meaningful reductions in storage size while keeping
the implementation efficient and straightforward.

\section{\texorpdfstring{Experiments }{Experiments }}\label{experiments}

In this section, we present empirical results demonstrating the
effectiveness of our lossless compression approach across multiple
settings and data types. We evaluate compression ratio, component-wise
breakdown (exponent vs. mantissa or scale), and consistency across
checkpoints or inference steps. All experiments were run on standard
transformer models with checkpoint data saved in BF16, FP8, or FP4
precision. Compression was applied per tensor using fixed-size chunks
and Huffman~\cite{huffman1952}-based entropy coding.

\subsection{Compression of BF16 Delta
Checkpoints}\label{compression-of-bf16-delta-checkpoints-1}

We begin with delta compression applied to full model checkpoints in
BF16 format. For each pair of consecutive checkpoints, we compute a
bitwise XOR to produce the delta. This delta is then split into exponent
and mantissa streams and compressed independently.

Figure 1 shows the compression ratios across four consecutive checkpoint
pairs. The exponent stream consistently shows strong compressibility.
The mantissa stream remains relatively high, between 0.69 and 0.92,
reflecting lower redundancy. Overall model compression reaches as low as
38 percent of the original delta size in later checkpoints, indicating
that redundancy increases as training converges.

These results confirm that delta compression in BF16 is highly effective
for lossless storage of intermediate model states. Most gains come from
the exponent, while the mantissa offers limited, though non-negligible,
savings.

\begin{quote}
\includegraphics{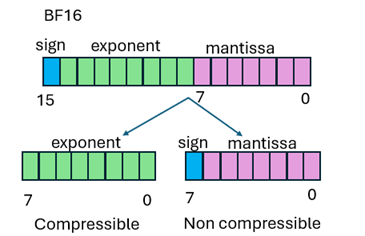}

\emph{Figure 5: Exponent Extraction for BF16 - group all exponents into
a separate compression stream}
\end{quote}

Dataset used is llm360/amber~\cite{llm360} which is an open-source data set of
checkpoints

\url{https://huggingface.co/LLM360/Amber}~\cite{llm360}

model size is 6.74B~params (and so each checkpoint size)

Compression granularity was done per checkpoint, per layer file, using
standard Huffman~\cite{huffman1952} encoding techniques.

\begin{quote}
\includegraphics{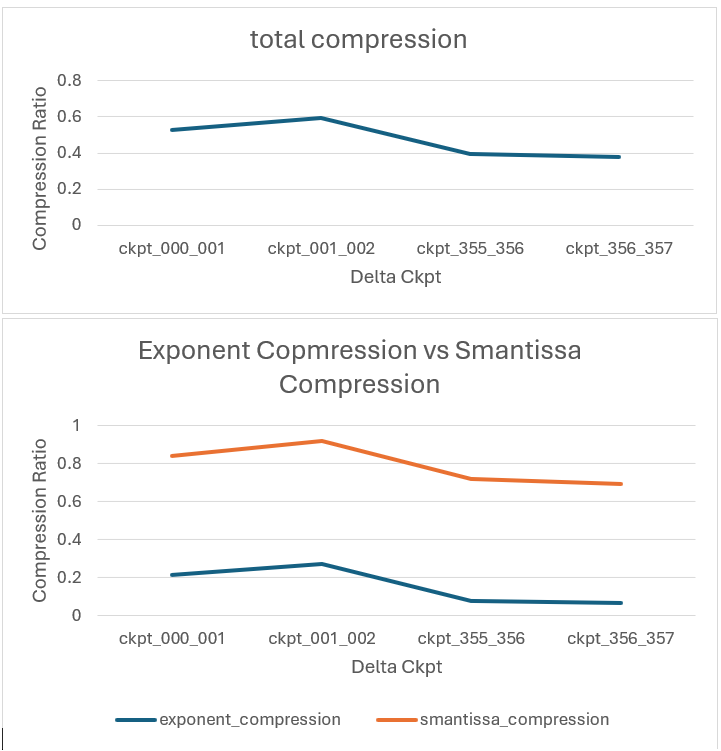}
Figure 6: compression ratios for delta compression of checkpoints on
amber data set~\cite{llm360}
\end{quote}

\subsection{Compression of FP8 Weights and
BF16}\label{compression-of-fp8-weights-and-bf16}

We evaluated our method on FP8 model weights using the E4M3 format
exclusively. This format was selected because its structure allows
straightforward byte alignment, which simplifies the processing of large
tensors and makes chunked compression more efficient.

For each tensor, we separated the exponent and mantissa bits. The
exponent values, though limited to 4 bits in E4M3, still showed skewed
distributions that were highly compressible. Across several transformer
layers, we observed exponent compression ratios ranging from 0.20 to
0.30. This reflects the tendency of neural network weights to occupy a
narrow dynamic range, even in low-precision representations.

The mantissa values, on the other hand, showed less redundancy and
therefore lower compressibility. Compression ratios for the mantissa
stream typically exceeded 0.80, with slight variations depending on
layer type and weight initialization. The total compression ratio for
the FP8 tensors ranged from 55 percent to 70 percent of the original
size, with the bulk of the savings attributed to the exponent component.

These results confirm that even in very low-bit formats like FP8,
separating the exponent from the mantissa allows us to exploit
statistical patterns that would be missed by generic compression tools.

\begin{quote}
\includegraphics[width=\linewidth]{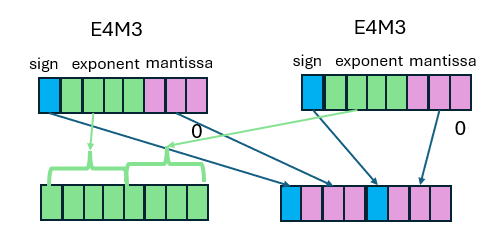}

\emph{Figure 7: grouping 2 elements of exponents, 2 elements of
sign+mantissa to easily store in a byte}
\end{quote}

Results:
\begin{quote}
\includegraphics[width=\textwidth]{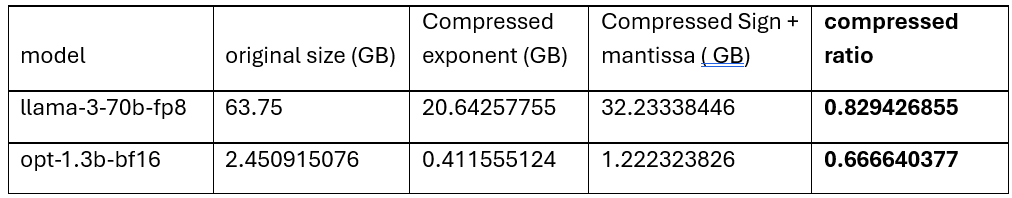}

\emph{Figure 8: Results of compression for FP8 and BF16 models
a separate compression stream}
\end{quote}

\subsection{Compression of K/V Cache
Tensors}\label{compression-of-kv-cache-tensors}

Key-value (K/V) cache tensors represent a significant source of memory
usage during inference in large language models. These tensors grow
proportionally with the number of generated tokens, the number of
layers, and the batch size. As sequence lengths increase, the memory
footprint of K/V caches can become a limiting factor for deployment,
particularly on resource-constrained devices.

Due to this challenge, several recent efforts have explored
architectural and system-level solutions. These include hierarchical
caching systems that move data between GPU and CPU memory, such as
HuggingFace's Offloaded Cache, and tiered storage systems
that extend caching to SSDs, such as NVIDIA's multi-level~\cite{nvidia_kvcache_blog} cache
framework. Other proposals, like RocketKV~\cite{rocketkv} and SpeCache~\cite{specache}, apply techniques
like eviction policies or speculative prefetching to manage memory usage
more effectively.

Our work complements these approaches by applying lossless compression
to the K/V cache tensors themselves. Specifically, we apply the same
exponent-mantissa separation used for model weights to K/V cache tensors
in FP8 and BF16 formats. Compression is applied per tensor, at each time
step during inference. We observed that exponent values, particularly in
early decoding steps, exhibit a strong concentration around a small set
of values. In FP8 caches, exponent compression ratios ranged from 0.25
to 0.45, while in BF16 the ratios were often below 0.20. Mantissa values
remained high-entropy and were stored without compression in most cases.

\subsection{Compression of FP4
Tensors}\label{compression-of-fp4-tensors}

FP4 formats are designed to maximize space efficiency by representing
each value with only 4 bits, typically as unsigned integers in the range
of 0 to 15. These formats rely on external scaling factors, stored in
higher-precision formats, to restore the full numerical range. We
evaluated compression for FP4 tensors under both MXFP4~\cite{ocp_mx_format} 
and NVFP4~\cite{nvidia_nvfp4}schemes.

In our experiments, we investigated whether the 4-bit data streams could
be compressed by exploiting potential bit-level structure. One approach
involved concatenating two bits from each of four consecutive FP4
elements to form a full byte. This allowed us to analyze the combined
stream using byte-aligned entropy coding. However, this strategy did not
yield useful compression. The resulting streams appeared statistically
uniform, with little to no redundancy across tensor blocks.

We then turned our attention to the scaling factors, which were stored
in FP8 format. These values showed strong compressibility due to
repeated patterns and limited dynamic variation across layers.

\begin{quote}
\includegraphics[width=0.7\textwidth]{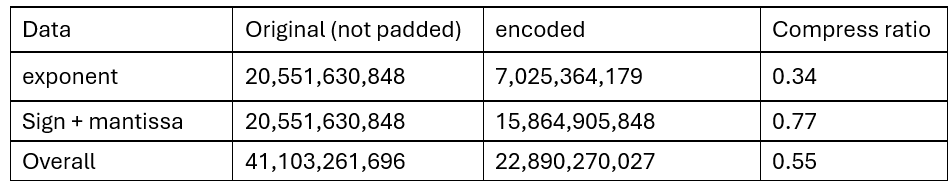}\\
Figure 9: Scalers of NVFP4 of DeepSeek-R1. There are 2 FP8 scalers for every 16 FP4 elements. The scalers size is 39 GB (out of original dataset of 395 GB), which means that overall compression for the dataset is 5\% as the scaling factors in NVFP4 composed 10\% of the overall dataset.
amber data set~\cite{llm360}
\end{quote}

\section{Discussion and future
directions}\label{discussion-and-future-directions}

The results presented in the previous section highlight several
important observations about the compressibility of low-precision
formats in neural networks. In this section, we analyze these outcomes,
compare the behavior across data types, and consider how our findings
relate to practical deployment scenarios.

\subsection{Compression Behavior Across
Formats}\label{compression-behavior-across-formats}

Our experiments confirm that exponent-mantissa separation, followed by
entropy-based encoding, remains a powerful tool for compressing neural
network tensors, even at reduced bit widths. In both BF16 and FP8
formats, the exponent field retains enough statistical skew to enable
substantial compression. This is particularly striking in FP8, where the
exponent is only 4 bits wide, yet consistently produces compression
ratios below 0.30. The mantissa component, by contrast, shows high
entropy and contributes little to overall savings unless sparsity or
rounding is involved.

In FP4, the situation is more constrained. The quantized 4-bit values
offer no useful compressibility due to their small range and uniform
distribution. However, the associated scaling factors, stored in FP8,
exhibit patterns that are well-suited for Huffman~\cite{huffman1952} coding. This
suggests that even in extremely compact formats, there are components
worth targeting for lossless compression.

Across all formats, exponent compression proves to be the most reliable
and impactful strategy. The separation of tensor components enables
focused application of lightweight algorithms, which is especially
important when aiming for high-speed or real-time inference scenarios.

\subsection{Value and Challenges of K/V Cache
Compression}\label{value-and-challenges-of-kv-cache-compression}

Among all the compression targets we explored, K/V cache tensors present
both the greatest opportunity and the greatest challenge. These tensors
store hidden states across layers and time steps during autoregressive
inference in large language models. Their memory usage scales with
sequence length, batch size, and number of layers, making them a major
bottleneck for long-context inference and large-batch deployments.

While model weights can be compressed offline and only need to be
decompressed once during execution, K/V cache tensors are generated and
accessed in real time. This introduces a significant constraint: both
encoding and decoding must occur on-the-fly, often within tight latency
budgets. As a result, applying even lightweight lossless compression to
K/V tensors requires high-throughput implementations that do not
interfere with the speed of token generation.

This real-time constraint differentiates K/V cache compression from most
other model compression scenarios. In practice, achieving usable
performance demands either aggressive software optimization or support
from dedicated hardware. Ideally, compression and decompression of the
exponent stream should be implemented as fast-path operations, possibly
with specialized accelerators or instructions that can operate in
parallel with model execution.

Despite these challenges, our results show that the exponent portion of
FP8 and BF16 K/V cache tensors are consistently compressible. When using
static Huffman~\cite{huffman1952} dictionaries trained on representative data, we
were able to reduce memory usage by 20 to 30 percent without introducing
significant overhead. These savings are compatible with and
complementary to existing industry efforts such as cache offloading,
hierarchical memory systems, and speculative cache management.

In the long term, further performance improvements in K/V cache
compression could help extend the context length of large models, reduce
the need for memory offloading, and make long-sequence inference more
efficient and accessible.

\section{Conclusion}\label{conclusion}

In this work, we extended lossless compression techniques for neural
network tensors to low-precision formats, with a particular focus on
FP8, FP4, and BF16. Building upon the principle of exponent-mantissa
separation, we demonstrated that even highly compact floating-point
representations retain sufficient statistical structure in their
exponent fields to enable meaningful compression using entropy-based
methods such as Huffman~\cite{huffman1952} coding.

Our method achieves significant compression for FP8 and BF16 weights,
with total size reductions of up to 45 percent in some cases. For FP4
tensors, we identified that only the scaling factors are compressible,
while the 4-bit quantized values offer little redundancy. These insights
suggest that targeting specific components of tensor representations is
essential for effective compression at low bit-widths.

A key contribution of this work is the application of lossless
compression to key-value cache tensors in large language models. We
showed that these tensors, despite being generated dynamically during
inference, exhibit compressible patterns in their exponent components.
This opens the door to real-time memory savings without introducing any
numerical changes to model behavior. However, the need for both encoding
and decoding to happen during live inference presents unique challenges.
Achieving low-latency compression for K/V caches will likely require
further optimizations, including hardware acceleration or integration
into inference runtimes.

Overall, our findings demonstrate that lossless compression remains
relevant and valuable even in the context of low-precision deployment.
These methods can serve as a complementary tool alongside quantization,
pruning, and caching strategies to improve model storage, transmission,
and runtime efficiency. Future work may explore compression-aware
training, adaptive runtime schemes, or extending these techniques to
other components such as activations, gradients, or optimizer states.

\bibliographystyle{unsrt}
\bibliography{references}
\end{document}